\documentclass[11pt,a4paper]{article}
\usepackage[hyperref]{acl2021}
\usepackage{times}
\usepackage{latexsym}

\usepackage{microtype}

\usepackage{amsmath}
\usepackage{url}

\usepackage{graphicx}
\usepackage{amsfonts}
\usepackage{CJK}
\usepackage{bm}
\usepackage{mathrsfs}
\usepackage{float}
\usepackage{xcolor,colortbl}

\usepackage{subfigure}
\usepackage{booktabs,multirow}
\usepackage{bigstrut,bigdelim}
\usepackage{paralist}
\usepackage{bbm}

\usepackage{helvet}
\usepackage{courier}
\usepackage{makecell}
\usepackage{nicefrac}
\usepackage{microtype}
\usepackage{epsfig}
\usepackage{diagbox}
\usepackage{framed}
\usepackage{empheq}
\usepackage{array}
\usepackage{enumitem}
\usepackage{mdwlist}
\usepackage{xspace,mfirstuc,tabulary}
\usepackage{tcolorbox}
\usepackage{algorithm}
\usepackage{placeins}
\usepackage{algpseudocode}

\usepackage{microtype}
\aclfinalcopy 
\def\aclpaperid{1881} 

\title{\textsc{MPC-BERT}: A Pre-Trained Language Model for Multi-Party Conversation Understanding}

\author{Jia-Chen Gu$^1$\thanks{\hspace{1.5mm}Work done during the internship at Microsoft.}, Chongyang Tao$^2$, Zhen-Hua Ling$^1$, Can Xu$^2$, Xiubo Geng$^2$, Daxin Jiang$^2$\thanks{\hspace{1.5mm}Corresponding author.} \\
  $^1$National Engineering Laboratory for Speech and Language Information Processing, \\
      University of Science and Technology of China, Hefei, China \\
  $^2$Microsoft, Beijing, China \\
{\tt gujc@mail.ustc.edu.cn}, {\tt zhling@ustc.edu.cn}, \\ {\tt \{chotao,caxu,xigeng,djiang\}@microsoft.com}
}

\date{}

\begin{document}
\maketitle
\begin{abstract}
  Recently, various neural models for multi-party conversation (MPC) have achieved impressive improvements on a variety of tasks such as addressee recognition, speaker identification and response prediction.
  However, these existing methods on MPC usually represent interlocutors and utterances individually and ignore the inherent complicated structure in MPC which may provide crucial interlocutor and utterance semantics and would enhance the conversation understanding process. 
  To this end, we present MPC-BERT, a pre-trained model for MPC understanding that considers learning \emph{who} says \emph{what} to \emph{whom} in a unified model with several elaborated self-supervised tasks.
  Particularly, these tasks can be generally categorized into (1) interlocutor structure modeling including {reply-to utterance recognition, identical speaker searching} and {pointer consistency distinction}, and (2) utterance semantics modeling including {masked shared utterance restoration} and {shared node detection}.
  We evaluate MPC-BERT on three downstream tasks including {addressee recognition, speaker identification} and {response selection}.
  Experimental results show that MPC-BERT outperforms previous methods by large margins and achieves new state-of-the-art performance on all three downstream tasks at two benchmarks.
\end{abstract}

\section{Introduction}
  Building a conversational agent with intelligence has drawn significant attention from both academia and industry.
  Most of existing methods have studied understanding conversations between two participants, aiming to return an appropriate response either in a generation-based \cite{DBLP:conf/acl/ShangLL15,DBLP:conf/aaai/SerbanSBCP16,DBLP:conf/aaai/SerbanSLCPCB17,DBLP:conf/nips/ZhangGGGLBD18,DBLP:conf/acl/ZhangSGCBGGLD20} or retrieval-based manner \cite{DBLP:conf/sigdial/LowePSP15,DBLP:conf/acl/WuWXZL17,DBLP:conf/acl/WuLCZDYZL18,DBLP:conf/wsdm/TaoWXHZY19,DBLP:conf/acl/TaoWXHZY19,DBLP:conf/cikm/GuLL19,DBLP:conf/emnlp/GuLZL19,DBLP:conf/cikm/GuLLLSWZ20}.
  Recently, researchers have paid more attention to a more practical and challenging scenario involving more than two participants, which is well known as multi-party conversation (MPC) \cite{DBLP:conf/emnlp/OuchiT16,DBLP:conf/aaai/ZhangLPR18,DBLP:conf/emnlp/LeHSYBZY19,DBLP:conf/ijcai/HuCL0MY19}.
  Table~\ref{tab-example} shows an MPC example in the Ubuntu Internet Relay Chat (IRC) channel, which is composed of a sequence of (\emph{speaker, utterance, addressee}) triples.
  In addition to returning an appropriate response, predicting who will be the next speaker \cite{DBLP:conf/lrec/MengMJ18} and who is the addressee of an utterance  \cite{DBLP:conf/emnlp/OuchiT16,DBLP:conf/aaai/ZhangLPR18,DBLP:conf/emnlp/LeHSYBZY19} are unique and important issues in MPC.

  \begin{table}[t]
    \small
    \centering
    \setlength{\tabcolsep}{2.4pt}
    \begin{tabular}{c|l|c}
      \toprule
        Speaker               &  \multicolumn{1}{c|}{Utterance}         &  Addressee \\
      \hline
        \multirow{2}{*}{I.1}  &  How can I setup if I want add new      &  \multirow{2}{*}{-}  \\
                              &  server at xchat? & \\
      \hline
        \multirow{3}{*}{I.2}  &  From places, network servers, work     &  \multirow{3}{*}{I.1} \\
                              &  group, his computer, and then I        &  \\
                              &  clicked on the shared folder.          &  \\
      \hline
        I.3                   &  It did not allow you to see the files? &  I.2 \\
      \hline
        \multirow{3}{*}{I.2}  &  It prompts for authentication and I    &  \multirow{3}{*}{I.3} \\
                              &  don't know what to put. I tried guest  &  \\
                              &  with no password.  \\
      \hline
        I.4                   &  Put proper authentication in, then?    &  I.2 \\
      \hline
        I.3                   &  I think you had kde on suse?           &  I.2 \\
      \bottomrule
      \end{tabular}
      \caption{An MPC example in Ubuntu IRC channel. Here, ``I."  is the abbreviation of  ``interlocutor".}
      \label{tab-example}
  \end{table}

  An instance of MPC always contains complicated interactions between interlocutors, between utterances and between an interlocutor and an utterance.
  Therefore, it is challenging to model the conversation flow and fully understand the dialogue content.
  Existing studies on MPC learn the representations of interlocutors and utterances with neural networks, and their representation spaces are either separate \cite{DBLP:conf/emnlp/OuchiT16} or interactive \cite{DBLP:conf/aaai/ZhangLPR18}. 
  However, the semantics contained in the interlocutor and utterance representations may not be effectively captured as they are from two different representation spaces.
  Recently, to take advantage of the breakthrough in pre-training language models (PLMs) for natural language understanding, some studies proposed to integrate the speaker \cite{DBLP:conf/cikm/GuLLLSWZ20} or topic \cite{DBLP:conf/emnlp/WangHJ20} information into PLMs. 
  Despite of the performance improvement on response selection, these models still overlook the inherent relationships between utterances and interlocutors, such as ``address-to". Furthermore, most existing studies design models for each individual task in MPC (e.g., addressee recognition, speaker identification and response prediction) separately.
  Intuitively, these tasks are complementary among each other. 
  Making use of these tasks simultaneously may produce better contextualized representations of interlocutors and utterances, and would enhance the conversation understanding, but is neglected in previous studies. 

  On account of above issues, we propose MPC-BERT which jointly learns \emph{who} says \emph{what} to \emph{whom} in MPC by designing self-supervised tasks for PLMs, so as to improve the ability of PLMs on MPC understanding.
  Specifically, 
  the five designed tasks includes \emph{reply-to utterance recognition, identical speaker searching, pointer consistency distinction, masked shared utterance restoration} and \emph{shared node detection}.
  The first three tasks are designed to model the interlocutor structure in MPC in a \emph{semantics-to-structure} manner.
  In the output of MPC-BERT, an interlocutor is described through the encoded representations of the utterances it says.
  Thus,  the representations of utterance semantics are utilized to construct the conversation structure in these three tasks.
  On the other hand, the last two tasks are designed to model the utterance semantics in a \emph{structure-to-semantics} manner.
  Intuitively, the conversation structure influences the information flow in MPC.
  Thus, the structure information can also be used to strengthen the representations of utterance semantics in return.
  In general, these five self-supervised tasks are employed to jointly train the MPC-BERT in a multi-task learning framework, which helps the model to learn the complementary information among interlocutors and utterances, and that between structure and semantics. 
  By this means, MPC-BERT can produce better interlocutor and utterance representations which can be effectively generalized to multiple downstream tasks of MPC.

  To measure the effectiveness of these self-supervised tasks and to test the generalization ability of MPC-BERT, we evaluate it on three downstream tasks including \emph{addressee recognition}, \emph{speaker identification} and \emph{response selection}, which are three core research issues of MPC. 
  Two benchmarks based on Ubuntu IRC channel are employed for evaluation.
  One was released by \citet{DBLP:conf/ijcai/HuCL0MY19}. 
  The other was released by \citet{DBLP:conf/emnlp/OuchiT16} and has three experimental settings according to session lengths.
  Experimental results show that MPC-BERT outperforms the current state-of-the-art models
  by margins of 3.51\%, 2.86\%, 3.28\% and 5.36\% on the test sets of these two benchmarks respectively in terms of the session accuracy of addressee recognition,
  by margins of 7.66\%, 2.60\%, 3.38\% and 4.24\% respectively in terms of the utterance precision of speaker identification, and
  by margins of 3.82\%, 2.71\%, 2.55\% and 3.22\% respectively in terms of the response recall of response selection.

  In summary, our contributions in this paper are three-fold: 
  (1) MPC-BERT, a PLM for MPC understanding, is proposed by designing five self-supervised tasks based on the interactions among utterances and interlocutors.
  (2) Three downstream tasks are employed to comprehensively evaluate the effectiveness of our designed self-supervised tasks and the generalization ability of MPC-BERT.
  (3) Our proposed MPC-BERT achieves new state-of-the-art performance on  all three downstream tasks at two benchmarks.
\section{Related Work}
  Existing methods on building dialogue systems can be generally categorized into studying two-party conversations and multi-party conversations (MPC). 
  In this paper, we study MPC. 
  In addition to predicting utterances, identifying the \emph{speaker} and recognizing the \emph{addressee} of an utterance are also important tasks for MPC.
  \citet{DBLP:conf/emnlp/OuchiT16} first proposed the task of addressee and response selection and created an MPC corpus for studying this task.
  \citet{DBLP:conf/aaai/ZhangLPR18} proposed SI-RNN, which updated speaker embeddings role-sensitively for addressee and response selection. 
  \citet{DBLP:conf/lrec/MengMJ18} proposed a task of speaker classification as a surrogate task for speaker modeling. 
  \citet{DBLP:conf/emnlp/LeHSYBZY19} proposed a who-to-whom (W2W) model to recognize the addressees of all utterances.
  \citet{DBLP:conf/ijcai/HuCL0MY19} proposed a graph-structured network (GSN) to model the graphical information flow for response generation.
  \citet{DBLP:conf/emnlp/WangHJ20} proposed to track the dynamic topic for response selection.

  Generally speaking, previous studies on MPC cannot unify the representations of interlocutors and utterances effectively.
  Also, they are limited to each individual task, ignoring the complementary information among different tasks.
  To the best of our knowledge, this paper makes the first attempt to design various self-supervised tasks for building PLMs aiming at MPC understanding, and to evaluate the performance of PLMs on three downstream tasks as comprehensively as possible.

\section{MPC-BERT and Self-Supervised Tasks}


  An MPC instance is composed of a sequence of (\emph{speaker, utterance, addressee}) triples, denoted as $\{(s_n,u_n,a_n)\}_{n=1}^N$, where $N$ is the number of turns in the conversation.
  Our goal is to build a pre-trained language model for universal MPC understanding.
  Given a conversation, this model is expected to produce embedding vectors for all utterances which contain not only the semantic information of each utterance, but also the speaker and addressee structure of the whole conversation.
  Thus, it can be effectively adapted to various downstream tasks by fine-tuning model parameters. 

  \subsection{Model Overview}
    In this paper, BERT \cite{DBLP:conf/naacl/DevlinCLT19} is chosen as the backbone of our PLM for MPC. 
    Thus, we name it MPC-BERT. 
    It is worth noting that our proposed self-supervised tasks for training MPC-BERT can also be applied to other types of PLMs.

    We first give an overview of the input representations and the overall architectures of MPC-BERT. 
    When constructing the input representations, in order to consider the speaker information of each utterance, \emph{speaker} embeddings \cite{DBLP:conf/cikm/GuLLLSWZ20} are introduced as shown in Figure~\ref{fig-pretrain-interlocutor}. 
    Considering that the set of interlocutors are inconsistent in different conversations, a position-based interlocutor embedding table is initialized randomly at first and updated during pre-training, 
    which means each interlocutor in a conversation is assigned with an embedding vector according to the order it appears in the conversation.
    Then, the speaker embeddings for each utterance can be derived by looking up this embedding table.
    The speaker embeddings are combined with standard token, position and segmentation embeddings and are then encoded by BERT. 
    The output embeddings of BERT corresponding to different input tokens are utilized by different self-supervised tasks for further calculation.

  \subsection{Tasks of Interlocutor Structure Modeling}

    \begin{figure*}[t]
      \centering
      \includegraphics[width=15cm]{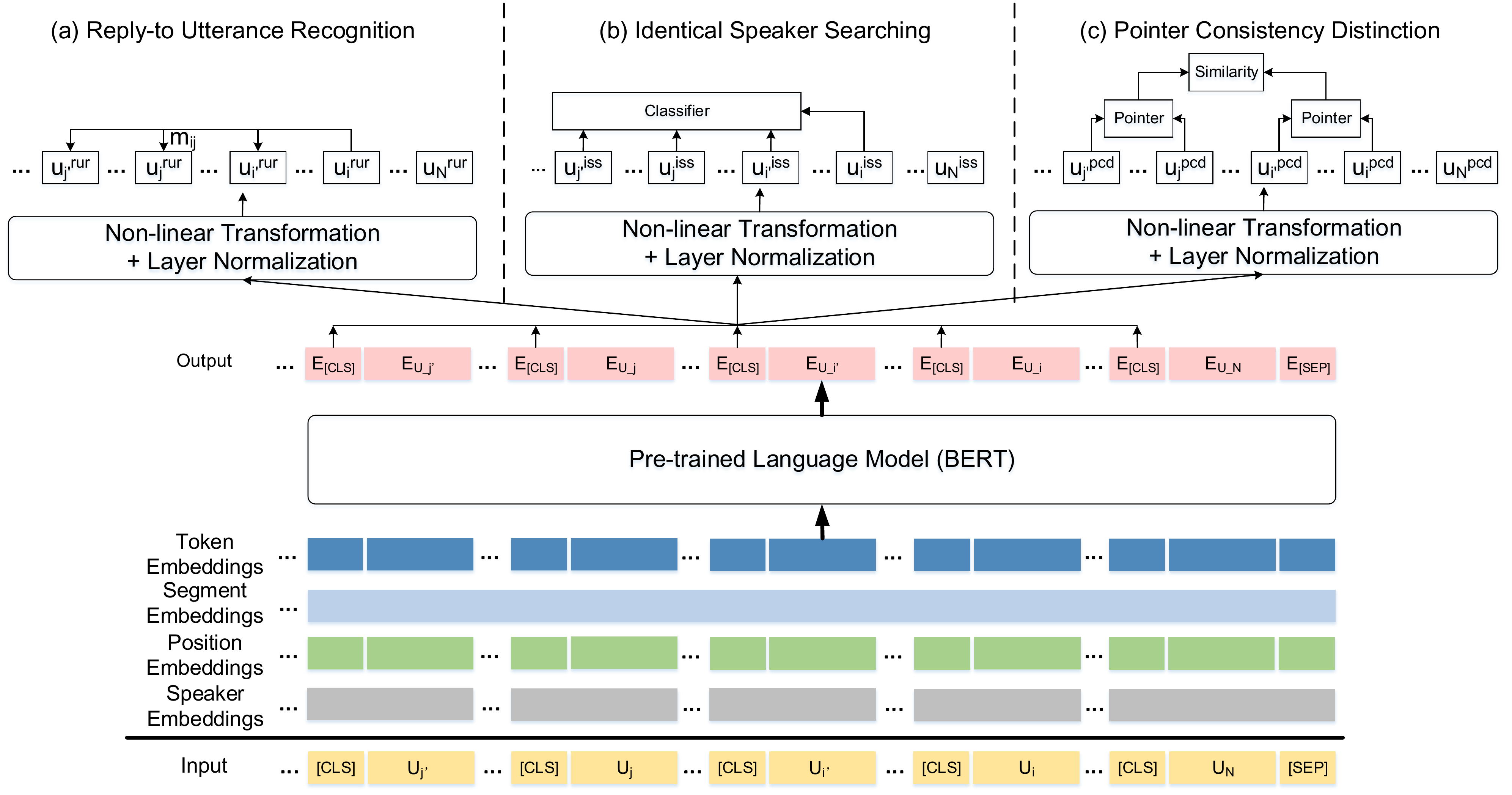}
      \caption{Input representations and model architectures of the three self-supervised tasks for interlocutor structure modeling, including (a) reply-to utterance recognition, (b) identical speaker searching and (c) pointer consistency distinction.
      }
      \label{fig-pretrain-interlocutor}
    \end{figure*}

    The first three tasks follow the \emph{semantics-to-structure} manner. 
    In MPC-BERT, each interlocutor is described through the encoded representations of the utterances it says. 
    Thus, the representations of utterance semantics are utilized to construct the conversation structure. 
    Figure~\ref{fig-pretrain-interlocutor} shows the input representations and the model architectures of these three tasks. 
    A \texttt{[CLS]} token is inserted at the start of each utterance, denoting its utterance-level representation. 
    Then, all utterances in a conversation are concatenated and a \texttt{[SEP]} token is inserted at the end of the whole sequence.
    It is notable that these three tasks share the same form of input data.
    Thus, the input only needs to be encoded once by BERT while the output can be fed into three tasks, which is computation-efficient.
    As shown in Figure~\ref{fig-pretrain-interlocutor}, a task-dependent non-linear transformation layer is placed on top of BERT in order to adapt the output of BERT to different tasks. 
    We will describe the details of these tasks as follows.

    \subsubsection{Reply-to Utterance Recognition}
      To enable the model to recognize the addressee of each utterance, a self-supervised task named \emph{reply-to utterance recognition (RUR)} is proposed to learn which preceding utterance the current utterance replies to.
      After encoded by BERT, we extract the contextualized representations for each \texttt{[CLS]} token representing individual utterances.
      Next, a non-linear transformation followed by a layer normalization are performed to derive the utterance representations for this specific task $\{\textbf{u}_i^{rur}\}_{i=1}^N$, where $\textbf{u}_i^{rur} \in \mathbb{R}^{d}$ and $d = 768$.
      Then, for a specific utterance U$_i$, its matching scores with all its preceding utterances are calculated as
      \begin{align}
        \label{equ-rur}
        m_{ij}  = \textbf{softmax} (\textbf{u}_i^{rur\top} \cdot \textbf{A}^{rur} \cdot \textbf{u}_j^{rur}),
      \end{align}
      where $\textbf{A}^{rur} \in \mathbb{R}^{d \times d}$ is a linear transformation, $m_{ij}$ denotes the matching degree of U$_j$ being the reply-to utterance of U$_i$, and $1\leq j<i$.
      We construct a set $\mathbb{S}$ by sampling a certain number of utterances in a conversation and this recognition operation is performed for each utterance in $\mathbb{S}$.
      Meanwhile, a dynamic sampling strategy is adopted so that models can see more samples.
      Finally, the pre-training objective of this self-supervised task is to minimize the cross-entropy loss as
      \begin{equation}
        \label{equ-rur-loss}
        \mathcal{L}_{rur} = - \sum_{i \in \mathbb{S}} \sum_{j=1}^{i-1} y_{ij} ~ log(m_{ij}),
      \end{equation}
      where $y_{ij} = 1$ if U$_j$ is the reply-to utterance of U$_i$ and  $y_{ij} = 0$ otherwise.

    \subsubsection{Identical Speaker Searching} \label{sec_iss}
      Having knowledge of who is the speaker of an utterance is also important for MPC.
      The task of \emph{identical speaker searching (ISS)} is designed by masking the speaker embedding of a specific utterance in the input representation, and aims to predict its speaker given the conversation.
      Since the set of interlocutors vary across conversations, the task of predicting the speaker of an utterance is reformulated as \emph{searching for the utterances sharing the identical speaker.}

      First, for a specific utterance, its speaker embedding is masked with a special \texttt{[Mask]} interlocutor embedding to avoid information leakage. 
      Given the utterance representations for this specific task $\{\textbf{u}_i^{iss}\}_{i=1}^N$ where $\textbf{u}_i^{iss} \in \mathbb{R}^{d}$, the matching scores of U$_i$ with all its preceding utterances are calculated similarly with Eq.~(\ref{equ-rur}).
      Here, $m_{ij}$ denotes the matching degree of U$_j$ sharing the same speaker with U$_i$.
      For each instance in the dynamic sampling set $\mathbb{S}$, there must be an utterance in previous turns sharing the same speaker. 
      Otherwise, it is removed out of the set. 
      Finally, the pre-training objective of this task is to minimize the cross-entropy loss similarly with Eq.~(\ref{equ-rur-loss}).
      Here, $y_{ij} = 1$ if U$_j$ shares the same speaker with U$_i$ and  $y_{ij} = 0$ otherwise.

    \subsubsection{Pointer Consistency Distinction}
\begin{figure}[t]
      \centering
      \subfigure[{Pointer consistency distinction}]{
      \includegraphics[width=0.45\linewidth]{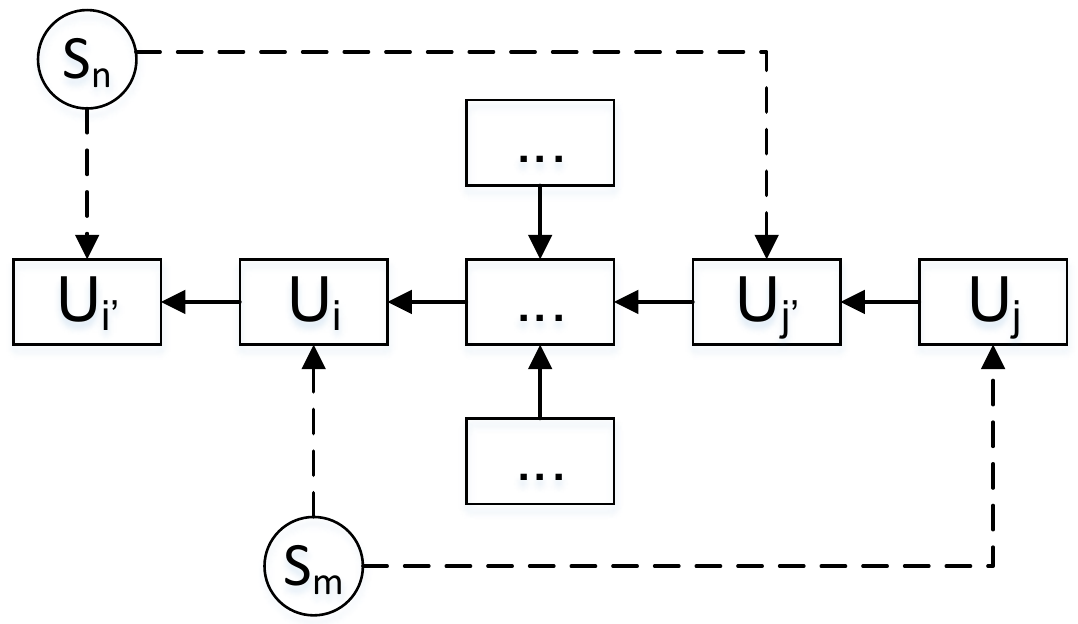}}
      \hspace{1.5mm}
      \subfigure[Shared node detection]{
      \includegraphics[width=0.45\linewidth]{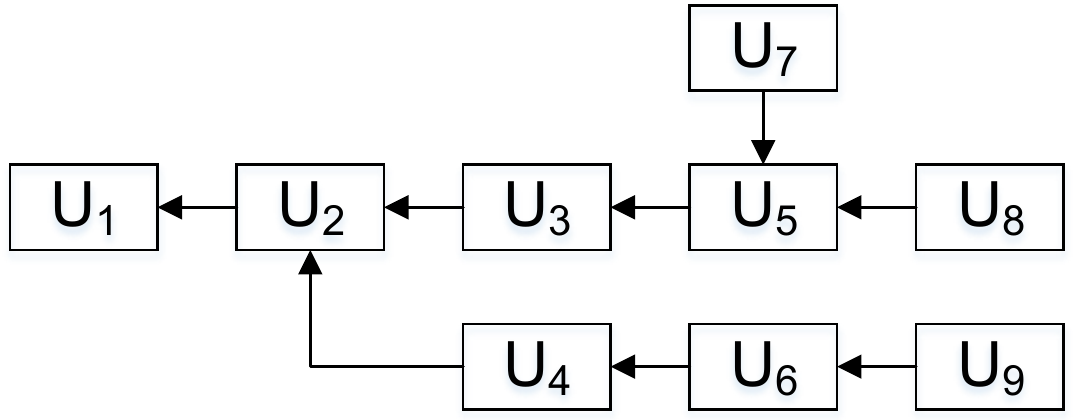}}
      \caption{Illustrations of the self-supervised tasks of (a) pointer consistency distinction and (b) shared node detection.
      Rectangles denote utterances, circles denote interlocutors, a solid line denotes an utterance replying to an utterance, and a dashed line denotes an utterance from an interlocutor.}
      \label{fig-pretrain-pcd-snd}
    \end{figure}

      We design a task named \emph{pointer consistency distinction (PCD)} to jointly model speakers and addressees in MPC.
      In this task, a pair of utterances representing the ``\emph{reply-to}'' relationship is defined as a \emph{speaker-to-addressee pointer}. 
      Here, we assume that the representations of two pointers directing from the same speaker to the same addressee should be consistent.
      As illustrated in Figure~\ref{fig-pretrain-pcd-snd} (a), speaker S$_m$ speaks U$_i$ and U$_j$ which reply to U$_{i'}$ and U$_{j'}$ from speaker S$_n$ respectively.
      Thus, the utterance tuples (U$_i$, U$_{i'}$) and (U$_j$, U$_{j'}$) both represent the pointer of S$_m$-to-S$_n$ and their pointer representations should be consistent..

      Given the utterance representations for this specific task $\{\textbf{u}_i^{pcd}\}_{i=1}^N$ where $\textbf{u}_i^{pcd} \in \mathbb{R}^{d}$, we first capture the pointer information contained in each utterance tuple.
      The element-wise difference and multiplication between an utterance tuple (U$_i$, U$_{i'}$) are computed and are concatenated as
      \begin{align}
        \textbf{p}_{ii'} = [\textbf{u}_i^{pcd} - \textbf{u}_{i'}^{pcd} ; \textbf{u}_i^{pcd} \odot \textbf{u}_{i'}^{pcd}],
      \end{align}
      where $\textbf{p}_{ii'} \in \mathbb{R}^{2d}$.
      Then, we compress $\textbf{p}_{ii'}$ and obtain the pointer representation $\bar{\textbf{p}}_{ii'}$ as
      \begin{align}
        \bar{\textbf{p}}_{ii'} = \textbf{ReLU}( \textbf{p}_{ii'} \cdot \textbf{W}_{pcd} + \textbf{b}_{pcd} ),
      \end{align}
      where $ \textbf{W}_{pcd} \in \mathbb{R}^{2d \times d} $ and $ \textbf{b}_{pcd} \in \mathbb{R}^{d} $ are parameters.
      Identically, a consistent pointer representations $\bar{\textbf{p}}_{jj'}$ and an inconsistent one $\bar{\textbf{p}}_{kk'}$ sampled from this conversation are obtained.
      The similarities between every two pointers are calculated as
      \begin{align}
        m_{ij}  = \textbf{sigmoid} (\bar{\textbf{p}}_{ii'}^{\top} \cdot \textbf{A}^{pcd} \cdot \bar{\textbf{p}}_{jj'}),
      \end{align}
      where $m_{ij}$ denotes the matching degree of pointer $\bar{\textbf{p}}_{ii'}$ being consistent with pointer $\bar{\textbf{p}}_{jj'}$.
      $m_{ik}$ can be derived accordingly.
      Finally, the pre-training objective of this task is to minimize the hinge loss which enforces $m_{ij}$ to be larger than $m_{ik}$ by at least a margin $\Delta$ as
      \begin{equation}
        \label{equ-pcd}
        \mathcal{L}_{pcd} = \textbf{max}\{ 0, \Delta - m_{ij} + m_{ik} \}.
      \end{equation}

  \subsection{Tasks of Utterance Semantics Modeling}
  
    Intuitively, the conversation structure might influence the information flow, so that it can be used to strengthen the representations of utterance semantics.
    Thus, two self-supervised tasks following the \emph{structure-to-semantics} manner are designed.

    \subsubsection{Masked Shared Utterance Restoration}
      There are usually several utterances replying-to a shared utterance in MPC. 
      Intuitively, a shared utterance is semantically relevant to more utterances in the context than non-shared ones. 
      Based on this characteristic, we design a task named \emph{masked shared utterance restoration (MSUR)}. 
      We first randomly sample an utterance from all shared utterances in a conversation and all tokens in this sampled utterance are masked with a \texttt{[MASK]} token.
      Then the model is enforced to restore the masked utterance given the rest conversation.

      Formally, assuming U$_i$ as the masked shared utterance and $l_i$ as the number of tokens in U$_i$.
      Given the token representations for this task  $\{\textbf{u}_{i,t}^{msur}\}_{t=1}^{l_i}$ where $\textbf{u}_{i,t}^{msur} \in \mathbb{R}^{d}$, the probability distribution of each masked token can be calculated as
      \begin{align}
        \textbf{p}_{u_{i,t}} = \textbf{softmax}( \textbf{u}_{i,t}^{msur} \cdot \textbf{W}_{msur} + \textbf{b}_{msur} ),
      \end{align}
      where $ \textbf{W}_{msur} \in \mathbb{R}^{d \times V} $ is the token embedding table, $V$ denotes the vocabulary size, and $\textbf{b}_{msur} \in \mathbb{R}^{V} $ is a bias vector.
      Finally, the pre-training objective of this self-supervised task is to minimize the negative log-likelihood loss as
      \begin{equation}
        \mathcal{L}_{msur} = - \frac{1}{l_i} \sum_{t=1}^{l_i} log~p_{u_{i,t}},
      \end{equation}
      where $p_{u_{i,t}}$ is the element in \textbf{p}$_{u_{i,t}}$ corresponding to the original token.

    \subsubsection{Shared Node Detection}
      A full MPC instance can be divided into several sub-conversations and
      we assume that the representations of sub-conversations under the same parent node tend to be similar.
      As illustrated in Figure~\ref{fig-pretrain-pcd-snd} (b), two sub-conversations \{U$_3$, U$_5$, U$_7$, U$_8$\} and \{U$_4$, U$_6$, U$_9$\} share the same parent node U$_2$. Thus, they should be semantically relevant.
      Under this assumption, we design a self-supervised task named \emph{shared node detection (SND)},
      which utilizes the conversation structure to strengthen the capability of models on measuring the semantic relevance of two sub-conversations.

      We first construct the pre-training samples for this task.
      Empirically, only the sub-conversations under the top shared node in a conversation are collected in order to filter out the sub-conversations with few utterances. 
      Given a full MPC, the two sub-conversations with the most utterances form a positive pair.
      For each positive pair, we replace one of its elements with another sub-conversation randomly sampled from the training corpus to form a negative pair. 

      Formally, given two sub-conversations $c_i$ and $c_j$, utterances in each sub-conversation are first concatenated respectively to form two segments.
      Then, the two segments are concatenated with a \texttt{[SEP]} token and a \texttt{[CLS]} token is inserted at the beginning of the whole sequence. 
      This sequence are encoded by BERT to derive the contextualized representation for the \texttt{[CLS]} token.
      A non-linear transformation with sigmoid activation is further applied to this representation for calculating the matching score $m_{ij}$, i.e., the probability of $c_i$ and $c_j$ sharing the same parent node.
      Finally, the pre-training objective of this task is to minimize the cross-entropy loss as
      \begin{equation}
        \label{equ-snd-loss}
        \mathcal{L}_{snd} =  - [ y_{ij} log(m_{ij}) + (1-y_{ij})log(1-m_{ij})],
      \end{equation}
      where $y_{ij} = 1$ if $c_i$ and $c_j$ share the same parent node and  $y_{ij} = 0$ otherwise.

  \subsection{Multi-task Learning}
    In addition, we also adopt the tasks of masked language model (MLM) and next sentence prediction (NSP) in original BERT pre-training \cite{DBLP:conf/naacl/DevlinCLT19}, which have been proven effective for incorporating  domain knowledge \cite{DBLP:conf/cikm/GuLLLSWZ20,DBLP:conf/acl/GururanganMSLBD20}.
    Finally, MPC-BERT is trained by performing multi-task learning that minimizes the sum of all loss functions as
    \begin{equation}
    \begin{aligned}
      \mathcal{L} = \mathcal{L}_{rur} & + \mathcal{L}_{iss} + \mathcal{L}_{pcd} + \mathcal{L}_{msur}  \\
                                      & + \mathcal{L}_{snd} + \mathcal{L}_{mlm} + \mathcal{L}_{nsp}.
    \end{aligned}
    \end{equation}

\section{Downstream Tasks}


  \subsection{Addressee Recognition}
    Given a multi-party conversation where part of the addressees are unknown, \citet{DBLP:conf/emnlp/OuchiT16} and \citet{DBLP:conf/aaai/ZhangLPR18} recognized an addressee of the last utterance. 
    \citet{DBLP:conf/emnlp/LeHSYBZY19} recognized addressees of all utterances in a conversation.
    In this paper, we follow the more challenging setting in \citet{DBLP:conf/emnlp/LeHSYBZY19}.

    Formally, models are asked to predict $\{\hat{a}_n\}_{n=1}^{N}$ given $\{(s_n,u_n,a_n)\}_{n=1}^N \backslash \{a_n\}_{n=1}^{N}$, where $\hat{a}_n$ is selected from the interlocutor set in this conversation and $\backslash$ denotes exclusion.
    When applying MPC-BERT, this task is reformulated as finding a preceding utterance from the same addressee. 
    Its RUR matching scores with all preceding utterances are calculated following Eq.~(\ref{equ-rur}).
    Then, the utterance with the highest score is selected and the speaker of the selected utterance is considered as the recognized addressee. 
    Finally, the fine-tuning objective of this task is to minimize the cross-entropy loss as
    \begin{equation}
      \mathcal{L}_{ar} = - \sum_{i=2}^N \sum_{j=1}^{i-1} y_{ij} ~ log(m_{ij}),
    \end{equation}
    where $m_{ij}$ is defined in Eq.~(\ref{equ-rur}), 
    $y_{ij} = 1$ if the speaker of U$_j$ is the addressee of U$_i$ and $y_{ij} = 0$ otherwise.

  \subsection{Speaker Identification}
    This task aims to identify the speaker of the last utterance in a conversation.
    Formally, models are asked to predict $\hat{s}_N$ given $\{(s_n,u_n,a_n)\}_{n=1}^N \backslash s_N$, where $\hat{s}_N$ is selected from the interlocutor set in this conversation. 
    When applying MPC-BERT, this task is reformulated as identifying the utterances sharing the same speaker.
    For the last utterance U$_N$, its speaker embedding is masked and its ISS matching scores $m_{Nj}$ with all preceding utterances are calculated following Section \ref{sec_iss}.
    The fine-tuning objective of this task is to minimize the cross-entropy loss as
    \begin{equation}
      \mathcal{L}_{si} = - \sum_{j=1}^{N-1} y_{Nj} ~ log(m_{Nj}),
    \end{equation}
    where 
    $y_{Nj} = 1$ if U$_j$ shares the same speaker with U$_N$ and $y_{Nj} = 0$ otherwise.

  \subsection{Response Selection}
    This task asks models to select $\hat{u}_N$ from a set of response candidates given the conversation context $\{(s_n,u_n,a_n)\}_{n=1}^N \backslash u_N$. 
    The key is to measure the similarity between two segments of context and response.
    We concatenate each response candidate with the context and extract the contextualized representation \textbf{e}$_{\texttt{[CLS]}}$ for the first \texttt{[CLS]} token using MPC-BERT.
    Then, \textbf{e}$_{\texttt{[CLS]}}$ is fed into a non-linear transformation with sigmoid activation to obtain the matching score between the context and the response. 
    Finally, the fine-tuning objective of this task is to minimize the cross-entropy loss according to the true/false labels of responses in the training set as
    \begin{equation}
      \mathcal{L}_{rs} =  - [ y log(m_{cr}) + (1-y)log(1-m_{cr})] ,
    \end{equation}
    where $y = 1$ if the response $r$ is a proper one for the context $c$; otherwise $y = 0$.

\section{Experiments}

    
    \begin{table}[t]
      \small
      \centering
      \setlength{\tabcolsep}{2.4pt}
      \begin{tabular}{c|c|c|c|c}
      \toprule
        \multicolumn{2}{c|}{Datasets}                                  &   Train   &   Valid   &  Test    \\
      \hline
        \multicolumn{2}{c|}{\citet{DBLP:conf/ijcai/HuCL0MY19}}         &  311,725  &  5,000    &  5,000   \\
      \hline
        \multirow{3}{*}{\citet{DBLP:conf/emnlp/OuchiT16}}   &  Len-5   &  461,120  &  28,570   &  32,668  \\
                                                            &  Len-10  &  495,226  &  30,974   &  35,638  \\
                                                            &  Len-15  &  489,812  &  30,815   &  35,385  \\
      \bottomrule
      \end{tabular}
      \caption{Statistics of the two benchmarks evaluated in this paper.}
      \label{tab-data}
    \end{table}
    

    \begin{table*}[t]
      \centering
      \setlength{\tabcolsep}{5.0pt}
      \begin{tabular}{l|c|c|c|c|c|c|c|c}
      \toprule
                                                     &  \multicolumn{2}{c|}{\citet{DBLP:conf/ijcai/HuCL0MY19}}  &  \multicolumn{6}{c}{\citet{DBLP:conf/emnlp/OuchiT16}}                         \\
      \cline{2-9}
                                                     &  \multicolumn{2}{c|}{}    &  \multicolumn{2}{c|}{Len-5}  &  \multicolumn{2}{c|}{Len-10}  &  \multicolumn{2}{c}{Len-15}  \\
      \cline{2-9}
                                                     &    P@1  &   Acc.             &    P@1  &   Acc.              &    P@1  &   Acc.             &   P@1   &   Acc.  \\
      \hline
        Preceding \cite{DBLP:conf/emnlp/LeHSYBZY19}  &    -    &    -               &  63.50  &  40.46             &  56.84  &  21.06              &  54.97  &  13.08  \\
        Subsequent \cite{DBLP:conf/emnlp/LeHSYBZY19} &    -    &    -               &  61.03  &  40.25             &  54.57  &  20.26              &  53.07  &  12.79  \\
        DRNN \cite{DBLP:conf/emnlp/OuchiT16}         &    -    &    -               &  72.75  &  58.18             &  65.58  &  34.47              &  62.60  &  22.58  \\
        SIRNN \cite{DBLP:conf/aaai/ZhangLPR18}       &    -    &    -               &  75.98  &  62.06             &  70.88  &  40.66              &  68.13  &  28.05  \\
        W2W \cite{DBLP:conf/emnlp/LeHSYBZY19}        &    -    &    -               &  77.55  &  63.81             &  73.52  &  44.14              &  73.42  &  34.23  \\
      \hline
        BERT \cite{DBLP:conf/naacl/DevlinCLT19}      &  96.16  &  83.50             &  85.95  &  75.99             &  83.41  &  58.22              &  81.09  &  44.94  \\
        SA-BERT \cite{DBLP:conf/cikm/GuLLLSWZ20}     &  97.12  &  88.91             &  86.81  &  77.45             &  84.46  &  60.30              &  82.84  &  47.23  \\
        MPC-BERT                                     &  \textbf{98.31}  &  \textbf{92.42}  & \textbf{88.73} & \textbf{80.31} & \textbf{86.23} & \textbf{63.58} & \textbf{85.55} & \textbf{52.59} \\
      \hline
        MPC-BERT w/o. RUR                            &  97.75  &  89.98             &  87.51  &  78.42             &  85.63  &  62.26              &  84.78  &  50.83  \\
        MPC-BERT w/o. ISS                            &  98.20  &  91.96             &  88.67  &  80.25             &  86.14  &  63.40              &  85.02  &  51.12  \\
        MPC-BERT w/o. PCD                            &  98.20  &  91.90             &  88.51  &  80.06             &  85.92  &  62.84              &  85.21  &  51.17  \\
        MPC-BERT w/o. MSUR                           &  98.08  &  91.32             &  88.70  &  80.26             &  86.21  &  63.46              &  85.28  &  51.23  \\
        MPC-BERT w/o. SND                            &  98.25  &  92.18             &  88.68  &  80.25             &  86.14  &  63.41              &  85.29  &  51.39  \\
      \bottomrule
      \end{tabular}
      \caption{Evaluation results of addressee recognition on the test sets. Results except ours are cited from \citet{DBLP:conf/emnlp/LeHSYBZY19}. Numbers in bold denote that the improvement over the best performing baseline is statistically significant (t-test with \emph{p}-value $<$ 0.05).}
      \label{tab-ar}
    \end{table*}


    \begin{table*}[t]
      \centering
      \begin{tabular}{l|c|c|c|c}
      \toprule
                                                     &  \multicolumn{1}{c|}{\citet{DBLP:conf/ijcai/HuCL0MY19}}  &  \multicolumn{3}{c}{\citet{DBLP:conf/emnlp/OuchiT16}}    \\
      \cline{2-5}
                                                     &         &  Len-5  &  Len-10  &  Len-15  \\
      \hline
        BERT \cite{DBLP:conf/naacl/DevlinCLT19}      &  71.81  &  62.24  &  53.17   &  51.58   \\
        SA-BERT \cite{DBLP:conf/cikm/GuLLLSWZ20}     &  75.88  &  64.96  &  57.62   &  54.28   \\
        MPC-BERT                                     &  \textbf{83.54}  & \textbf{67.56} & \textbf{61.00} & \textbf{58.52}  \\
      \hline
        MPC-BERT w/o. RUR                            &  82.48  &  66.88  &   60.12  &  57.33   \\
        MPC-BERT w/o. ISS                            &  77.95  &  66.77  &   60.03  &  56.73   \\
        MPC-BERT w/o. PCD                            &  83.39  &  67.12  &   60.62  &  58.00   \\
        MPC-BERT w/o. MSUR                           &  83.51  &  67.21  &   60.76  &  58.03   \\
        MPC-BERT w/o. SND                            &  83.47  &  67.04  &   60.44  &  58.12   \\
      \bottomrule
      \end{tabular}
      \caption{Evaluation results of speaker identification on the test sets in terms of P@1. Numbers in bold denote that the improvement over the best performing baseline is statistically significant (t-test with \emph{p}-value $<$ 0.05).}
      \label{tab-si}
    \end{table*}


    \begin{table*}[t]
      \centering
      \setlength{\tabcolsep}{4.0pt}
      \begin{tabular}{l|c|c|c|c|c|c|c|c}
      \toprule
                                                 &  \multicolumn{2}{c|}{\citet{DBLP:conf/ijcai/HuCL0MY19}}  &  \multicolumn{6}{c}{\citet{DBLP:conf/emnlp/OuchiT16}}  \\
      \cline{2-9}
                                                 &  \multicolumn{2}{c|}{}     & \multicolumn{2}{c|}{Len-5} &  \multicolumn{2}{c|}{Len-10}  &  \multicolumn{2}{c}{Len-15}  \\
      \cline{2-9}
                                                 & R$_2@1$ & R$_{10}@1$       & R$_2@1$ & R$_{10}@1$       & R$_2@1$ & R$_{10}@1$          & R$_2@1$ & R$_{10}@1$   \\
      \hline
        DRNN \cite{DBLP:conf/emnlp/OuchiT16}     &   -     &   -              &  76.07  &   33.62          &  78.16  &   36.14             &  78.64  &   36.93      \\
        SIRNN \cite{DBLP:conf/aaai/ZhangLPR18}   &   -     &   -              &  78.14  &   36.45          &  80.34  &   39.20             &  80.91  &   40.83      \\
      \hline
        BERT \cite{DBLP:conf/naacl/DevlinCLT19}  &  92.48  &  73.42           &  85.52  &   53.95          &  86.93  &   57.41             &  87.19  &   58.92      \\
        SA-BERT \cite{DBLP:conf/cikm/GuLLLSWZ20} &  92.98  &  75.16           &  86.53  &   55.24          &  87.98  &   59.27             &  88.34  &   60.42      \\
        MPC-BERT                                 &  \textbf{94.90}  &  \textbf{78.98}  & \textbf{87.63} & \textbf{57.95} & \textbf{89.14} & \textbf{61.82} & \textbf{89.70} & \textbf{63.64}  \\
      \hline
        MPC-BERT w/o. RUR                        &  94.48  &  78.16           &  87.20  &   57.56          &  88.96  &   61.47             &  89.07  &   63.24      \\
        MPC-BERT w/o. ISS                        &  94.58  &  78.82           &  87.54  &   57.77          &  88.98  &   61.76             &  89.58  &   63.51      \\
        MPC-BERT w/o. PCD                        &  94.66  &  78.70           &  87.50  &   57.51          &  88.75  &   61.62             &  89.45  &   63.46      \\
        MPC-BERT w/o. MSUR                       &  94.36  &  78.22           &  87.11  &   57.58          &  88.59  &   61.05             &  89.25  &   63.20      \\
        MPC-BERT w/o. SND                        &  93.92  &  76.96           &  87.30  &   57.54          &  88.77  &   61.54             &  89.27  &   63.34      \\
      \bottomrule
      \end{tabular}
      \caption{Evaluation results of response selection on the test sets. Results except ours are cited from \citet{DBLP:conf/emnlp/OuchiT16} and \citet{DBLP:conf/aaai/ZhangLPR18}. Numbers in bold denote that the improvement over the best performing baseline is statistically significant (t-test with \emph{p}-value $<$ 0.05).}
      \label{tab-rs}
    \end{table*}
    

  \subsection{Datasets}
    We evaluated our proposed methods on two Ubuntu IRC benchmarks.
    One was released by \citet{DBLP:conf/ijcai/HuCL0MY19}, in which both speaker and addressee labels was provided for each utterance.
    The other benchmark was released by \citet{DBLP:conf/emnlp/OuchiT16}.
    Here, we adopted the version shared in \citet{DBLP:conf/emnlp/LeHSYBZY19} for fair comparison. 
    The conversation sessions were separated into three categories according to the session length (Len-5, Len-10 and Len-15) following the splitting strategy of previous studies \cite{DBLP:conf/emnlp/OuchiT16,DBLP:conf/aaai/ZhangLPR18,DBLP:conf/emnlp/LeHSYBZY19}.
    Table~\ref{tab-data} presents the statistics of the two benchmarks evaluated in our experiments.

  \subsection{Baseline Models}
    \paragraph{Non-pre-training-based models} \citet{DBLP:conf/emnlp/OuchiT16} proposed a dynamic model DRNN which updated speaker embeddings with the conversation flow.
    \citet{DBLP:conf/aaai/ZhangLPR18} improved DRNN to SI-RNN which updated  speaker embeddings role-sensitively.
    \citet{DBLP:conf/emnlp/LeHSYBZY19} proposed W2W which jointly modeled interlocutors and utterances in a uniform framework, and predicted all addressees.

    \paragraph{Pre-training-based models} BERT \cite{DBLP:conf/naacl/DevlinCLT19} was pre-trained to learn general language representations with MLM and NSP tasks. 
    SA-BERT \cite{DBLP:conf/cikm/GuLLLSWZ20} added speaker embeddings and further pre-trained BERT on a domain-specific corpus to incorporate domain knowledge. 
    We re-implemented SA-BERT with the pre-training corpus used in this paper to ensure fair comparison.
    
    \begin{figure*}[t]
      \centering
      \subfigure[Addressee recognition]{
      \includegraphics[width=5cm]{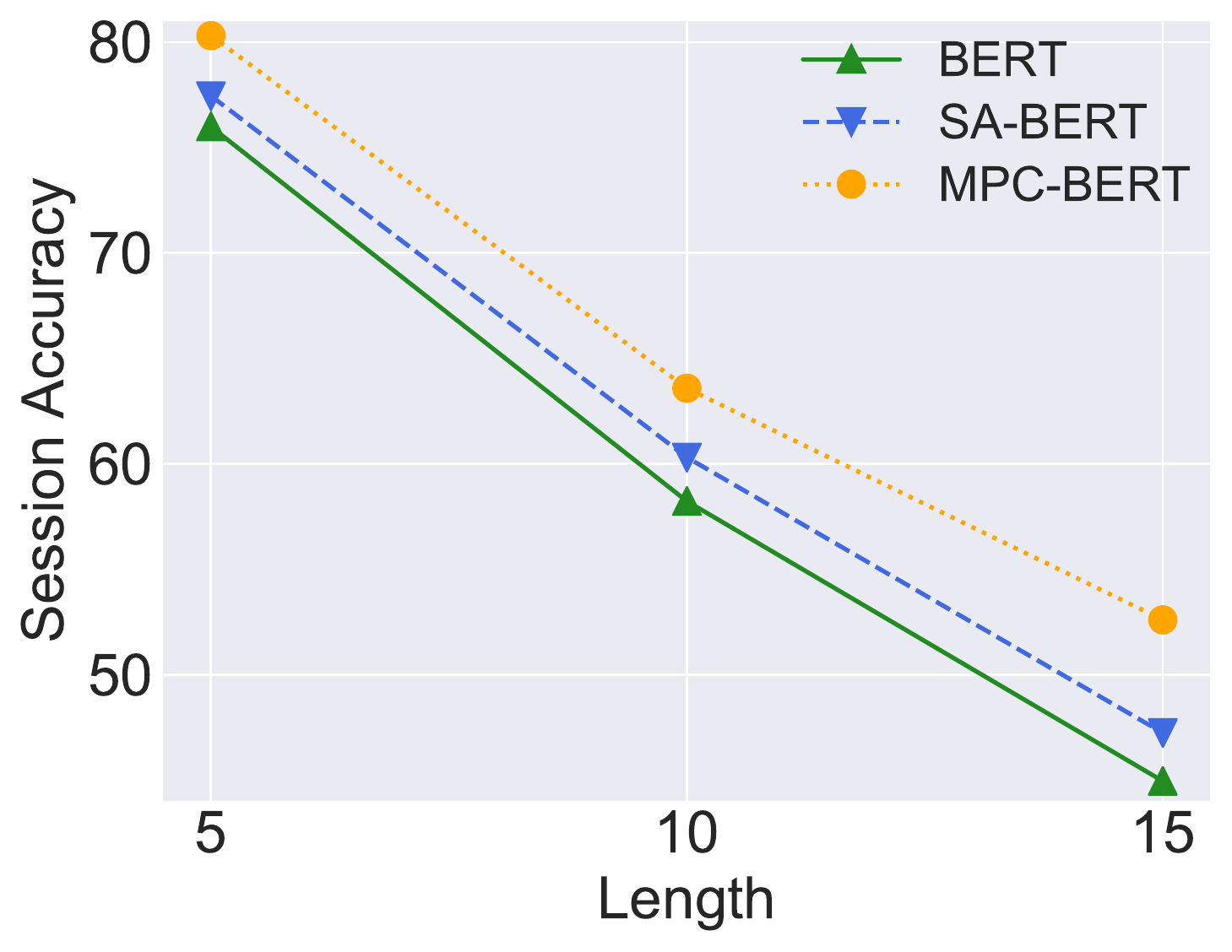}}
      \subfigure[Speaker identification]{
      \includegraphics[width=5cm]{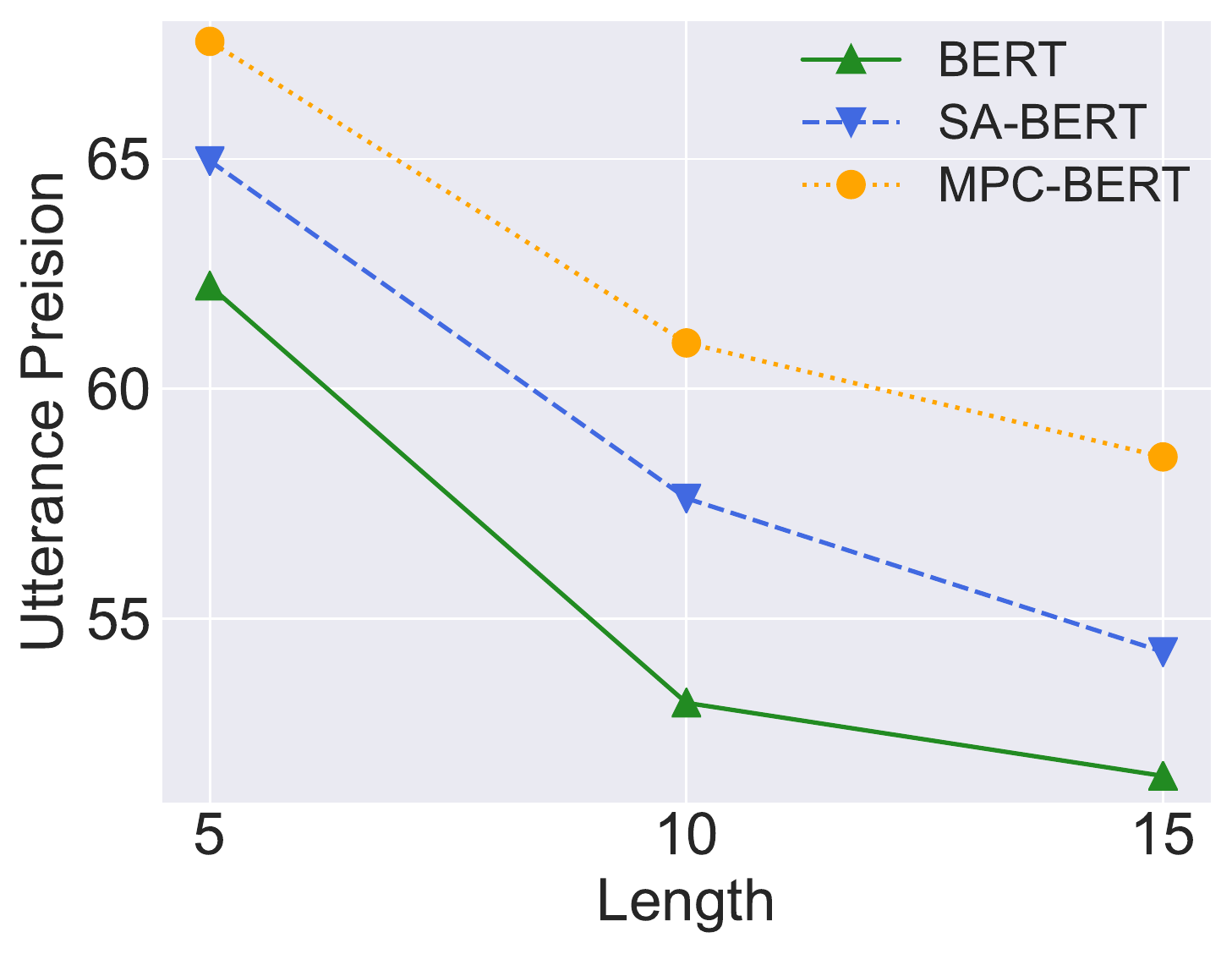}}
      \subfigure[Response selection]{
      \includegraphics[width=5cm]{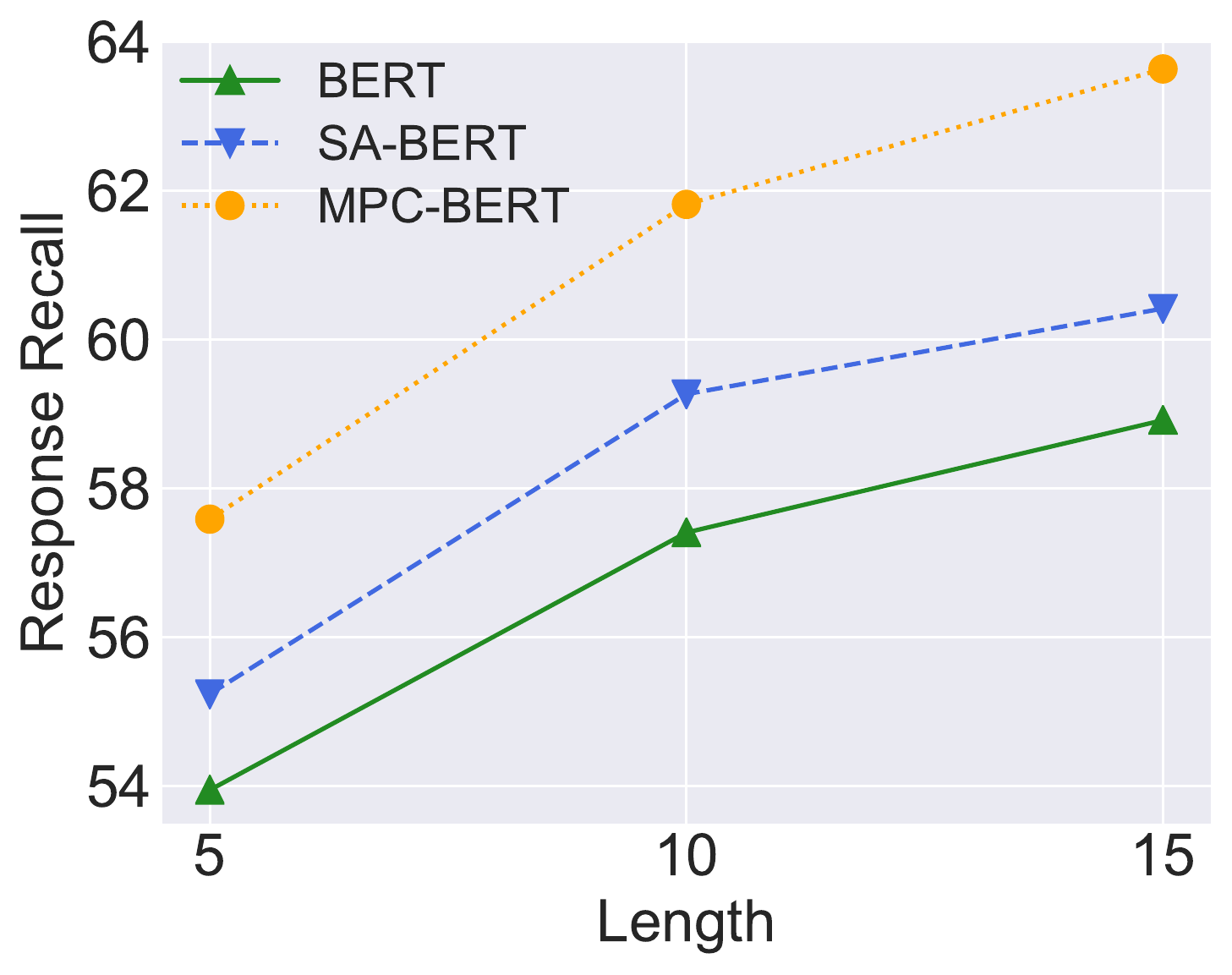}}
      \caption{Performance of models under different session lengths on the test sets of \citet{DBLP:conf/emnlp/OuchiT16} on the tasks of (a) addressee recognition, (b) speaker identification and (c) response selection.
      }
      \label{fig-analysis}
    \end{figure*}

  \subsection{Implementation Details} \label{sec-details}
    The version of BERT-base-uncased was adopted for all our experiments.
    For pre-training, \texttt{GELU} \cite{DBLP:journals/corr/HendrycksG16} was employed as the activation for all non-linear transformations.
    The Adam method \cite{DBLP:journals/corr/KingmaB14} was employed for optimization.
    The learning rate was initialized as 0.00005 and the warmup proportion was set to 0.1.
    We pre-trained BERT for 10 epochs. 
    The training set of the dateset used in \citet{DBLP:conf/ijcai/HuCL0MY19} was employed for pre-training. 
    The maximum utterance number was set to 7. 
    The maximum sequence length was set to 230.
    The maximum sampling numbers for each example were set to 4 for RUR, 2 for ISS and 2 for PCD.
    $\Delta$ in Eq.~(\ref{equ-pcd}) was set to 0.4, achieving the best performance out of \{0.2, 0.4, 0.6, 0.8\} on the validation set.
    The pre-training was performed using a GeForce RTX 2080 Ti GPU and the batch size was set to 4. 

    For fine-tuning, some configurations were different according to the characteristics of these datasets.
    For \citet{DBLP:conf/ijcai/HuCL0MY19}, the maximum utterance number was set to 7 and the maximum sequence length was set to 230.
    For the three experimental settings in \citet{DBLP:conf/emnlp/OuchiT16}, the maximum utterance numbers were set to 5, 10 and 15, and the maximum sequence lengths were set to 120, 220 and 320.
    All parameters in PLMs were updated.
    The learning rate was initialized as 0.00002 and the warmup proportion was set to 0.1.
    For \citet{DBLP:conf/ijcai/HuCL0MY19}, the fine-tuning process was performed for 10 epochs for addressee recognition, 10 epochs for speaker identification, and 5 epochs for response selection.
    For \citet{DBLP:conf/emnlp/OuchiT16}, the fine-tuning epochs were set to 5, 5 and 3 respectively. 
    The fine-tuning was also performed using a GeForce RTX 2080 Ti GPU. 
    The batch sizes were set to 16 for \citet{DBLP:conf/ijcai/HuCL0MY19}, and 40, 20, and 12 for the three experimental settings in \citet{DBLP:conf/emnlp/OuchiT16} respectively.
    The validation set was used to select the best model for testing.

    All codes were implemented in the TensorFlow framework \cite{DBLP:conf/osdi/AbadiBCCDDDGIIK16} and are published to help replicate our results.~\footnote{https://github.com/JasonForJoy/MPC-BERT}

  \subsection{Metrics and Results}

    \paragraph{Addressee recognition}
    We followed the metrics of previous work \cite{DBLP:conf/emnlp/LeHSYBZY19} by employing precision@1 (P@1) to evaluate each utterance with ground truth. 
    Also, a session is marked as positive if the addressees of all its utterances are correctly recognized, which is calculated as accuracy (Acc.).

    Table~\ref{tab-ar} presents the results of addressee recognition.
    It shows that MPC-BERT outperforms the best performing model, i.e., SA-BERT, by margins of 3.51\%, 2.86\%, 3.28\% and 5.36\% on these test sets respectively in terms of Acc., verifying the effectiveness of the proposed five self-supervised tasks as a whole.
    To further illustrate the effectiveness of each task, ablation tests were performed as shown in the last five rows of Table~\ref{tab-ar}.
    We can observe that all self-supervised tasks are useful as removing any of them causes performance drop. 
    Among the five tasks, RUR plays the most important role,
    and the tasks focusing on modeling interlocutor structure contribute more than those for utterance semantics.

    \paragraph{Speaker identification}
    Similarly, P@1 was employed as the evaluation metric of speaker identification for the last utterance of a conversation and the results are shown in  Table~\ref{tab-si}. 
    It shows that MPC-BERT outperforms SA-BERT by margins of 7.66\%, 2.60\%, 3.38\% and 4.24\% respectively in terms of P@1. 
    Besides, from the ablation results we find that all tasks are useful for improving the performance of speaker identification and ISS and RUR contribute the most. 
    In particular, removing PCD, MSUR and SND only leads to slight performance drop. The reason might be that the information conveyed by these tasks is redundant. 

    \paragraph{Response selection}
    The R$_n@k$ metrics adopted by previous studies  \cite{DBLP:conf/emnlp/OuchiT16,DBLP:conf/aaai/ZhangLPR18} were used here. 
    Each model was tasked with selecting $k$ best-matched responses from $n$ available candidates, and we calculated the recall as R$_n@k$. 
    Two settings were followed in which \emph{k} was set to 1 and \emph{n} was set to 2 or 10.

    Table~\ref{tab-rs} presents the results of response selection.
    It shows that MPC-BERT outperforms SA-BERT by margins of 3.82\%, 2.71\%, 2.55\% and 3.22\% respectively in terms of R$_{10}@1$. 
    Ablation tests show that SND is the most useful task for response selection and the two tasks focusing on the utterance semantics contribute more than those focusing on the interlocutor structures.

  \subsection{Discussions}
    Figure~\ref{fig-analysis} illustrates how the performance of BERT, SA-BERT and MPC-BERT changed with respect to different session lengths on the test sets of \citet{DBLP:conf/emnlp/OuchiT16}.
    It can be seen that the performance of addressee recognition and speaker identification dropped as the session length increased. 
    The reason might be that longer sessions always contain more interlocutors which increase the difficulties of predicting interlocutors.
    Meanwhile, the performance of response selection was significantly improved as the session length increased.
    It can be attributed to that longer sessions enrich the representations of contexts with more details which benefit response selection.
    Furthermore, as the session length increased, the performance of MPC-BERT dropped more slightly than that of SA-BERT on addressee recognition and speaker identification, and the R$_{10}@1$ gap between MPC-BERT and SA-BERT on response selection enlarged from 2.71\% to 3.22\%.
    These results imply the superiority of MPC-BERT over SA-BERT on modeling long MPCs with complicated structures.

\section{Conclusion}
  In this paper, we present MPC-BERT, a pre-trained language model with five self-supervised tasks for MPC understanding. 
  These tasks jointly learn \emph{who} says \emph{what} to \emph{whom} in MPCs.
  Experimental results on three downstream tasks show that MPC-BERT outperforms previous methods by large margins and achieves new state-of-the-art performance on two benchmarks.

\section*{Acknowledgments}
  We thank anonymous reviewers for their valuable comments.

\bibliographystyle{acl_natbib}
\bibliography{acl2021}


\end{document}